# Distribution of complexities in the Vai script

*Andrij Rovenchak[1], Lviv*
*Ján Mačutek[2], Bratislava*
*Charles Riley[3], New Haven, Connecticut*

**Abstract.** In the paper, we analyze the distribution of complexities in the Vai script, an indigenous syllabic writing system from Liberia. It is found that the uniformity hypothesis for complexities fails for this script. The models using Poisson distribution for the number of components and hyper-Poisson distribution for connections provide good fits in the case of the Vai script.

## 1. Introduction

Our study concentrates mainly on the complexity of the Vai script. We use the composition method suggested by Altmann (2004). It has some drawbacks (e. g., as mentioned by Köhler 2008, letter components are not weighted by their lengths, hence a short straight line in the letter G contributes to the letter complexity by 2 points, the same score is attributed to each of four longer lines of the letter M), but they are overshadowed by several important advantages (it is applicable to all scripts, it can be done relatively easily without a special software). And, of course, there is no perfect method in empirical science. Some alternative methods are mentioned in Altmann (2008).

Applying the Altmann's composition method, a letter is decomposed into its components (points with complexity 1, straight lines with complexity 2, arches not exceeding 180 degrees with complexity 3, filled areas[4] with complexity 2) and connections (continuous with complexity 1, crisp with complexity 2, crossing with complexity 3). Then, the letter complexity is the sum of its components and connections complexities. E. g., the letter O is assigned complexity 8 (2 arches, 2 continuous connections), the letter X has complexity 7 (2 straight lines, 1 crossing). See Altmann (2004) and Mačutek (2008) for a more detailed discussion on the method. In some cases the method is not unambiguous, e. g., sometimes a researcher must decide if he considers a thick line or a filled area with its contours. Different fonts of the same script usually yield different complexities.

From among scripts so far analyzed with respect to complexity we mention Latin (fonts Arial and Courier New, Altmann 2004), Cyrillic (its Ukrainian version, Buk, Mačutek and Rovenchak 2008) and several types of runes (Mačutek 2008).

---

[1] Department for Theoretical Physics, Ivan Franko National University of Lviv, 12 Drahomanov St., Lviv, UA-79005, Ukraine, e-mail: andrij@ktf.franko.lviv.ua, andrij.rovenchak@gmail.com

[2] Department of Applied Mathematics and Statistics, Comenius University, Mlynská dolina, 824 48 Bratislava, Slovakia, e-mail: jmacutek@yahoo.com

[3] Sterling Memorial Library, Yale University, 120 High Street, New Haven, CT 06511, USA, e-mail: charles.riley@yale.edu

[4] Altmann (2004) proposed complexity 1 for filled areas.

## 2. The geographic range of the Vai people and their language

Vai (also Vei, Vy, Gallinas, Gallines, phonetically [vaɪ]) is a Western Mande language belonging to the Niger-Congo language family. It is spoken by some 144,000 people, of which about 122,000 live in Liberia and some 22,000 in Sierra Leone[5]. The territory of the Vai speakers is shown in the map (Fig. 1). It is located on the Atlantic coast and stretches from the Lake Mabesi in the West to the Lofa River in the East, its northern boundary lying some ten miles south from the city of Potoru in Sierra Leone (roughly, this territory lies between 11°40′W and 11°00′W, and below 7°20′N). Note, that the Sierra Leone part is to a large extent shared with other peoples, namely, Mende (belonging to the Mande group) and Gola (speaking a language from the Atlantic group).

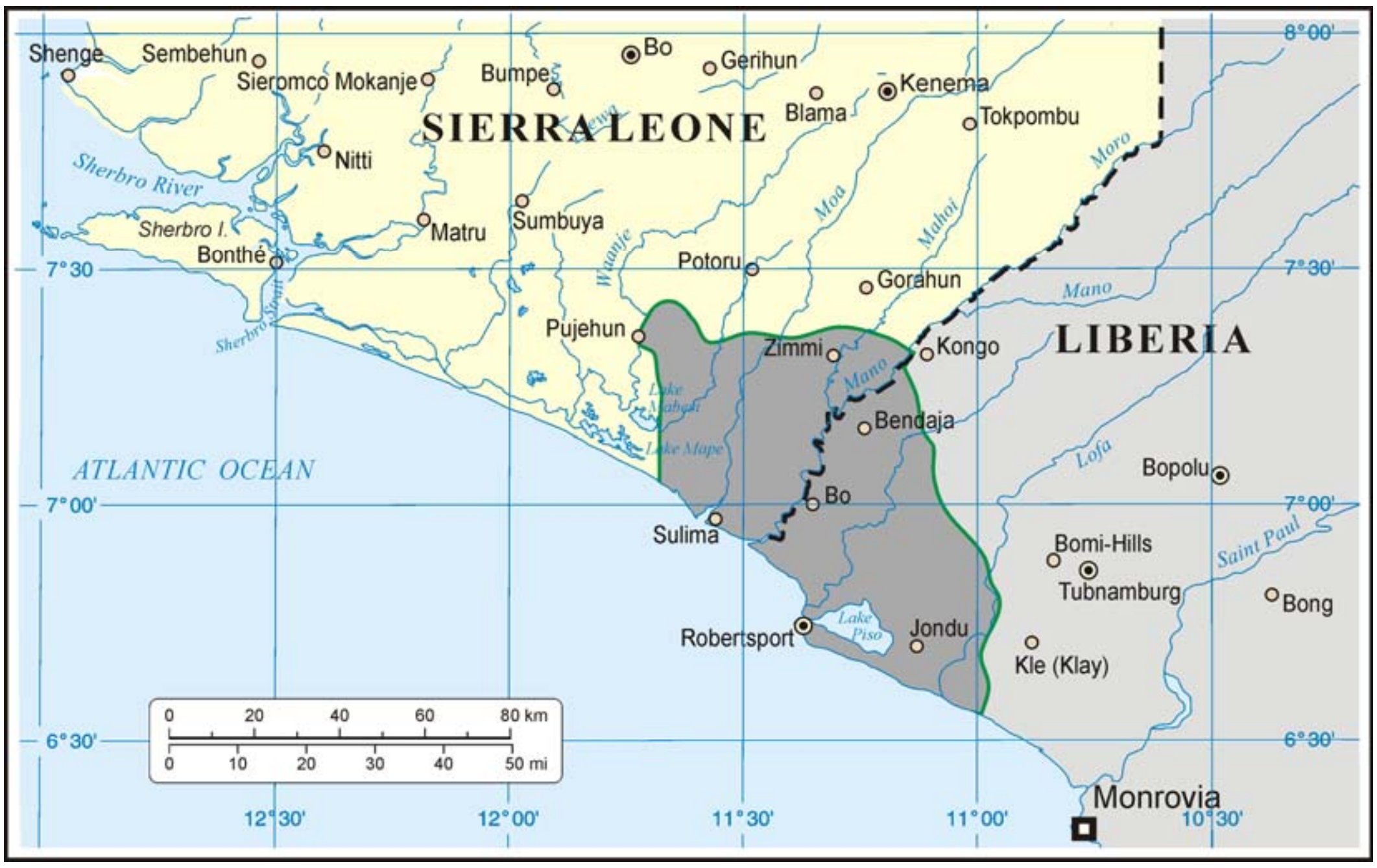

Fig. 1: Map showing the location of the Vai country (dark shaded area).

The Vai language has a remarkable phonology, reflecting its lexical history (Welmers, 1976). There are seven oral vowels [a, ɛ, e, i, ɔ, o, u], five nasal vowels[6] [ã, ɛ̃, ĩ, ɔ̃, ũ], and 31 consonant [b, ɓ, ʧ, d, ɗ, f, g, gb, h, ʤ, k, kp, l, m, mɓ, mgb, n, nɗ, ɲ, ɲʤ, ŋ, ŋg, p, s, t, v, w, j, z, r, ʃ]. The consonants [r, ʃ] are found only in relatively recent loans (John Singler, personal communication, 2005), and [ʧ] is probably relatively recent as well (Ofri-Scheps 1991).

[5] There is a large uncertainty about these numbers. The most recent available data can be found in Gordon (2005). These are 89,500 in Liberia and 15,500 in Sierra Leone for 1991. The results of recently conducted 2008 Census in Liberia with regard to ethnicity are not known yet. To make the estimation of the population, we took the population growth in Liberia between 1984 and 2008 (Census 2008). We are grateful to Prof. William Kory for the discussion on this issue. It must be noted that the number of the Vais in Sierra Leone defined by primary language use during the 2004 Census is significantly lower, about 2,500 (Thekeka Conteh, personal communication on July 10, 2008), when comparing to the estimated number of 22,000 obtained from the population growth in Sierra Leone.

[6] Note that the status of [ũ] is unclear: it is represented by a single syllabic sign hũ, probably an obsolete one (Priest 2004) and not included in the vowels list by modern authors (Welmers 1976, Ofri-Scheps 1991). Ofri-Scheps excludes [ĩ] as well, which is elsewhere represented by a single sign hĩ.

### 3. Vai script: organization, history, usage

The Vai script was created in 1820–30s in Jondu, Cape Mount, Liberia. Traditionally, Mɔmɔlu Duwalu Bukɛlɛ (?–1850) is credited as an inventor of the script. He presumably was assisted in this work by his five friends (Dalby 1967). It appears possible that the emergence of the Vai script is linked with the "stimulus diffusion" from the Cherokee syllabic script. Despite large distance from the Cherokees living in Oklahoma and the Vais in Liberia, this writing system could be known from the American mission in early 1820s: one of the missionaries was Austin Curtis, the Cherokee himself and at the same time one of the leading men in the Vai country (Tuchscherer & Hair 2002; Mafundikwa 2004; Tuchscherer 2005; 2007). The Vai script was widely used in the private sphere, where it survived until today. According to some data, every fifth Vai man can write it (Singler 1983; Tuchscherer 2005; 2007). The translations of Qur'an and Bible in the Vai script are also known. It is assumed that the Vai script became a direct stimulus for the creation of several other indigenous writing systems in the Western Africa, in particular Mende, Loma, Kpelle (Dalby 1967), and Bambara (Galtier 1987).

The Vai script is a syllabary. Not all the combination of the above-listed sounds are equally possible, thus the total number of syllables is far below the available arithmetically calculated maximum of some 400. The Vai syllabary counts over 200 signs, of which seven stand for individual oral vowels, two can be treated as independent nasals [ã, ɛ̃], remaining signs denoting open syllables plus one sign for syllabic [ŋ] (Dalby 1967; Jensen 1969; Coulmas 2004), see Table 1. Since the creation, many Vai signs changed their appearance, and the syllabary was standardized in 1899 and finally in 1962 by a committee in the University of Liberia (Singler 1996: 594). It is worth to note that in 1911 Momolu Massaquoi made an attempt to fill in the gaps of the original syllabary modifying the signs by adding diacritical marks (Massaquoi 1911). Some Massaquoi additions survived until today, in particular his signs for r- and ʃ-series; some were abandoned by Vai practitioners.

While Vai is a tonal language, the tones are not marked in the script. Such deficiency is however common even for many Roman orthographies used in African languages (Bird 1999). Of indigenous African writing systems, only in Bassa are the tones marked in a systematic way (Coulmas 2004), and – with a much higher precision – in the N'ko alphabet (Vydrine 1999).

In fact, "the basic unit of the system is more accurately the mora" (Singler 1996: 594). Syllables containing a long vowel are written with two signs, the second one often belonging to the h-series. The velar nasal [ŋ] is treated as a separate unit. Surprisingly, similar situation with mora-to-sign correspondence is found in, e. g., Japanese *kana*.

The Vais prefer European punctuation and digits, however, there exist native Vai punctuation signs (Jensen 1969, Massaquoi 1911). The Vai digits are also known but they are not widely used (Everson et al. 2006), these are modified European digits to conform the Vai style.

The direction of the Vai script is from left to right in horizontal lines going from top to bottom.

### 4. Some quantitative analyses

Table 1 contains the list of the Vai syllabic signs based mainly on the Dukor typeface (courtesy of Evertype). This font reflects the style given in Tucker (1999). Recently, the Vai script became a part of the Unicode standard, version 5.1. This fact also helps to standardize the shape of individual signs, which varies a lot in handwritten texts.

Table 1

Complexity of Vai letters

| | sign | transliteration | components | connections | complexity |
|---|---|---|---|---|---|
| 1 | | a | 4×1+1×2+2×3 | 2×1+1×2 | 16 |
| 2 | | ɛ | 6×3 | 5×1+3×3 | 32 |
| 3 | | e | 3×2+4×3 | 4×1+2×2 | 26 |
| 4 | | i | 4×2+2×3 | 5×2 | 24 |
| 5 | | ɔ | 6×3 | 5×2 | 28 |
| 6 | | o | 6×2 | 5×2 | 22 |
| 7 | | u | 2×1+3×2+2×3 | 5×2 | 24 |
| 8 | | ã | 3×3 | 2×1+1×3 | 14 |
| 9 | | ɛ̃ | 1×2+5×3 | 2×1+4×2 | 27 |
| 10 | | ba | 4×3 | 3×1+2×3 | 21 |
| 11 | | bɛ | 2×1+4×2 | 3×2 | 16 |
| 12 | | be | 2×1+2×2+6×3 | 6×1+3×2 | 36 |
| 13 | | bi | 6×3 | 4×1+1×2+2×3 | 30 |
| 14 | | bɔ | 1×2+8×3 | 6×1+2×2+2×3 | 42 |
| 15 | | bo | 2×1+2×2+4×3 | 3×1+2×2 | 25 |
| 16 | | bu | 6×3 | 6×1 | 24 |
| 17 | | ɓa | 1×1+4×2+1×3 | 6×2 | 24 |
| 18 | | ɓɛ | 1×2+1×3 | — | 5 |
| 19 | | ɓe | 2×2+1×3 | 2×2 | 11 |
| 20 | | ɓi | 2×2+4×3 | 4×1+4×2 | 28 |
| 21 | | ɓɔ | 3×2+1×3+1×2* | 4×2 | 19 |
| 22 | | ɓo | 1×1+5×2 | 4×2 | 19 |
| 23 | | ɓu | 5×3 | 4×1+5×2 | 25 |
| 24 | | ʧa | 2×2+2×3 | 1×1+3×2 | 17 |
| 25 | | ʧɛ | 2×2+4×3 | 2×1+3×2 | 24 |
| 26 | | ʧe | 5×3+2×2 | 4×1+2×3 | 29 |
| 27 | | ʧi | 2×2+2×3 | 1×1+3×2 | 17 |
| 28 | | ʧɔ | 2×1+2×2+2×3 | 6×2 | 24 |
| 29 | | ʧo | 4×1+1×2 | — | 6 |

| | sign | transliteration | components | connections | complexity |
|---|---|---|---|---|---|
| 30 | 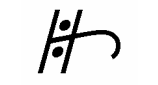 | tʃu | 2×1+2×2+1×3 | 1×2+1×3 | 14 |
| 31 | 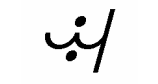 | da | 2×1+1×2+2×3 | 2×2 | 14 |
| 32 | 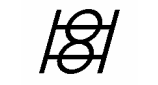 | dɛ | 4×2+4×3 | 4×1+4×2+4×3 | 46 |
| 33 | 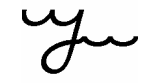 | de | 7×3 | 2×1+4×2+1×3 | 34 |
| 34 | 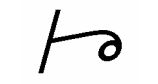 | di | 1×2+3×3 | 2×1+2×2 | 17 |
| 35 | 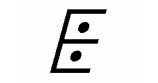 | dɔ | 2×1+4×2 | 3×2 | 16 |
| 36 | 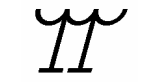 | do | 4×2+3×3 | 8×2 | 33 |
| 37 | 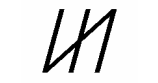 | du | 4×2 | 2×2+1×3 | 15 |
| 38 | 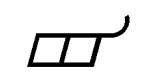 | ɗa | 4×2+1×3 | 6×2 | 23 |
| 39 | 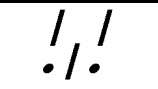 | ɗɛ | 2×1+3×2 | — | 8 |
| 40 | 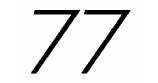 | ɗe | 4×2 | 2×2 | 12 |
| 41 | 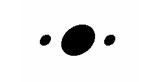 | ɗi | 2×1+2×3+1×2 | 2×1 | 12 |
| 42 | 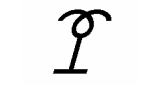 | ɗɔ | 2×2+3×3 | 2×1+2×2+1×3 | 22 |
| 43 | 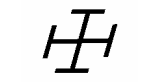 | ɗo | 6×2 | 4×2+1×3 | 23 |
| 44 | 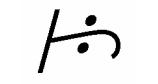 | ɗu | 2×1+1×2+1×3 | 1×2 | 9 |
| 45 | 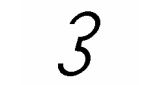 | fa | 1×2+4×3 | 2×1+2×2 | 20 |
| 46 | 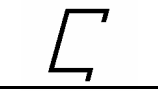 | fɛ | 5×2 | 4×2 | 18 |
| 47 | 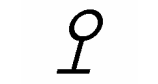 | fe | 2×2+2×3 | 2×1+2×2 | 16 |
| 48 | 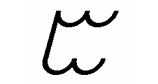 | fi | 4×3 | 3×2 | 18 |
| 49 | 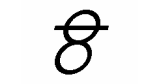 | fɔ | 1×2+4×3 | 4×1+1×2+2×3 | 26 |
| 50 | 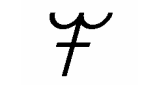 | fo | 2×2+2×3 | 3×2+1×3 | 19 |
| 51 | 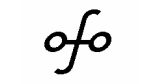 | fu | 1×2+6×3 | 5×1+2×2+1×3 | 32 |
| 52 | 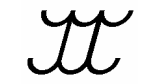 | ga | 6×3 | 8×2 | 34 |
| 53 | 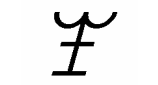 | gɛ | 3×2+2×3 | 4×2+1×3 | 23 |
| 54 | 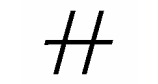 | ge | 3×2 | 2×3 | 12 |
| 55 | 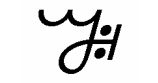 | gi | 2×1+1×2+4×3 | 1×1+3×2+1×3 | 26 |
| 56 | 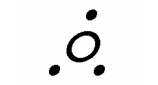 | gɔ | 3×1+2×3 | 2×1 | 11 |
| 57 | 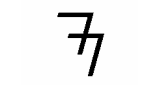 | go | 4×2 | 2×2+1×3 | 15 |
| 58 | 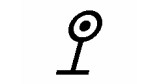 | gu | 1×1+2×2+2×3 | 2×1+2×2 | 17 |
| 59 | 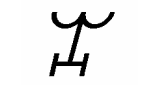 | gɛ̃ | 4×2+2×3 | 6×2 | 26 |
| 60 | 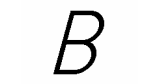 | gba | 1×2+2×3 | 5×2 | 18 |
| 61 | 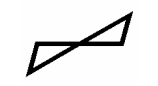 | gbɛ | 4×2 | 4×2+1×3 | 19 |
| 62 | 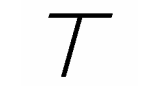 | gbe | 2×2 | 1×2 | 6 |
| 63 | 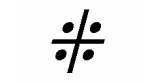 | gbi | 4×1+2×2 | 1×3 | 11 |
| 64 | 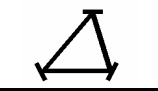 | gbɔ | 6×2 | 9×2 | 30 |
| 65 | 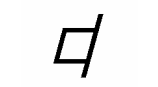 | gbo | 4×2 | 4×2 | 16 |
| 66 | 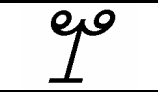 | gbu | 2×2+6×3 | 4×1+6×2 | 38 |
| 67 | 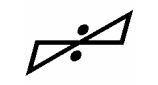 | gbɛ̃ | 2×1+4×2 | 4×2+1×3 | 21 |
| 68 | 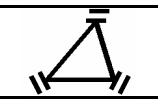 | gbɔ̃ | 9×2 | 9×2 | 36 |
| 69 | 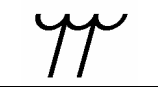 | ha | 2×2+3×3 | 6×2 | 25 |

|  | sign | transliteration | components | connections | complexity |
|---|---|---|---|---|---|
| 70 | 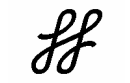 | hɛ | 8×3 | 7×1+4×3 | 43 |
| 71 | 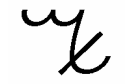 | he | 1×2+3×3 | 2×2+1×3 | 18 |
| 72 | 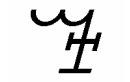 | hi | 5×2+2×3 | 5×2+1×3 | 29 |
| 73 | 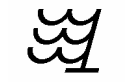 | hɔ | 2×2+6×3 | 7×2 | 36 |
| 74 | 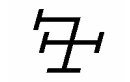 | ho | 7×2 | 5×2+1×3 | 27 |
| 75 | 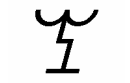 | hu | 4×2+2×3 | 6×2 | 26 |
| 76 | 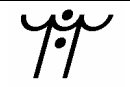 | hã | 2×1+2×2+3×3 | 6×2 | 27 |
| 77 | 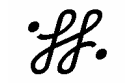 | hɛ̃ | 2×1+8×3 | 7×1+4×3 | 45 |
| 78 | 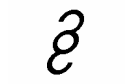 | hĩ | 6×3 | 4×1+7×2 | 26 |
| 79 | 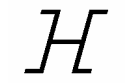 | hɔ̃ | 7×2 | 6×2 | 26 |
| 80 | 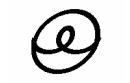 | hũ | 5×3 | 4×1+3×3 | 28 |
| 81 | 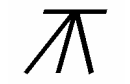 | ʤa | 4×2 | 6×2 | 20 |
| 82 | 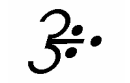 | ʤɛ | 3×1+2×2+4×3 | 2×1+3×2 | 27 |
| 83 | 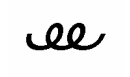 | ʤe | 5×3 | 4×1+2×3 | 25 |
| 84 |  | ʤi | 3×3 | 2×2 | 13 |
| 85 | 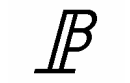 | ʤɔ | 3×2+2×3 | 8×2 | 28 |
| 86 | 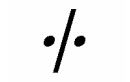 | ʤo | 2×1+1×2 | — | 4 |
| 87 | 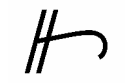 | ʤu | 2×2+1×3 | 1×2+1×3 | 12 |
| 88 | 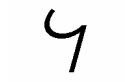 | ka | 1×2+1×3 | 1×2 | 7 |
| 89 | 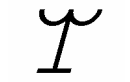 | kɛ | 2×2+2×3 | 4×2 | 18 |
| 90 | 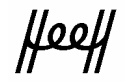 | ke | 4×2+4×3 | 3×1+2×2+2×3 | 39 |
| 91 | 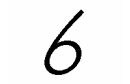 | ki | 2×3 | 1×1+1×2 | 9 |
| 92 | 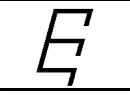 | kɔ | 7×2 | 6×2 | 26 |
| 93 | 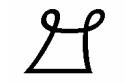 | ko | 1×2+3×3 | 2×1+2×2+2×3 | 23 |
| 94 | 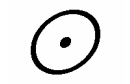 | ku | 1×1+2×3 | 2×1 | 9 |
| 95 | 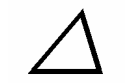 | kpa | 3×2 | 3×2 | 12 |
| 96 |  | kpɛ | 1×2+4×3 | 4×1+2×2 | 22 |
| 97 | 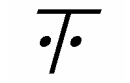 | kpe | 2×1+2×2 | 1×2 | 8 |
| 98 | 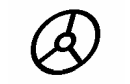 | kpi | 3×2+4×3 | 4×1+6×2 | 34 |
| 99 | 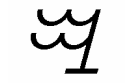 | kpɔ | 2×2+4×3 | 5×2 | 26 |
| 100 | 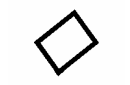 | kpo | 4×2 | 4×2 | 16 |
| 101 | 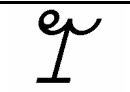 | kpu | 2×2+4×3 | 2×1+5×2 | 28 |
| 102 | 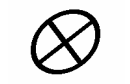 | kpã | 2×2+2×3 | 2×1+4×2+1×3 | 23 |
| 103 | 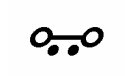 | kpɛ̃ | 2×1+1×2+4×3 | 4×1+2×2 | 24 |
| 104 | 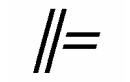 | la | 4×2 | — | 8 |
| 105 | 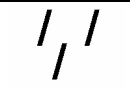 | lɛ | 3×2 | — | 6 |
| 106 | 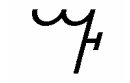 | le | 3×2+2×3 | 4×2 | 20 |
| 107 | 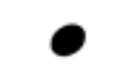 | li | 2×3+1×2 | 2×1 | 10 |
| 108 | 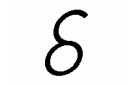 | lɔ | 3×3 | 2×1+1×2 | 13 |
| 109 | 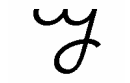 | lo | 4×3 | 1×1+2×2+1×3 | 20 |

| | sign | transliteration | components | connections | complexity |
|---|---|---|---|---|---|
| 110 | 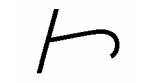 | lu | 1×2+1×3 | 1×2 | 7 |
| 111 | 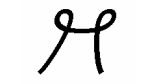 | ma | 3×3 | 2×1+2×3 | 17 |
| 112 | 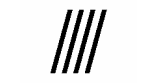 | mɛ | 4×2 | — | 8 |
| 113 | 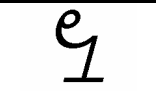 | me | 2×2+3×3 | 2×1+3×2 | 21 |
| 114 | 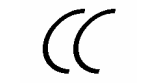 | mi | 2×3 | — | 6 |
| 115 | 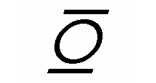 | mɔ | 2×2+2×3 | 2×1 | 12 |
| 116 | 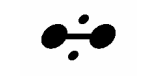 | mo | 2×1+1×2+4×3+2×2 | 4×1+2×2 | 28 |
| 117 | 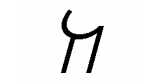 | mu | 2×2+1×3 | 2×2 | 11 |
| 118 | 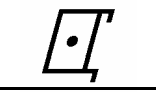 | mɓa | 1×1+6×2 | 6×2 | 25 |
| 119 | 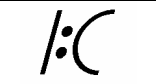 | mɓɛ | 2×1+1×2+1×3 | — | 7 |
| 120 | 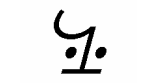 | mɓe | 2×1+2×2+1×3 | 2×2 | 13 |
| 121 | 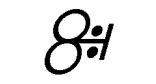 | mɓi | 2×1+2×2+4×3 | 4×1+4×2 | 30 |
| 122 | 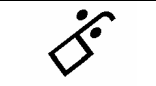 | mɓɔ | 2×1+3×2+1×3 | 4×2 | 19 |
| 123 |  | mɓo | 1×2+4×3+2×2 | 4×1+2×2 | 26 |
| 124 | 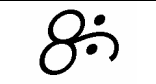 | mɓu | 2×1+5×3 | 4×1+3×2 | 27 |
| 125 | 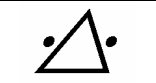 | mgba | 2×1+3×2 | 3×2 | 14 |
| 126 | 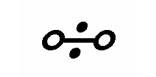 | mgbɛ | 2×1+1×2+4×3 | 4×1+2×2 | 24 |
| 127 | 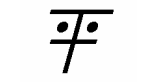 | mgbe | 2×1+3×2 | 1×2+1×3 | 13 |
| 128 | 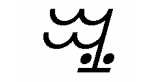 | mgbɔ | 2×1+2×2+4×3 | 5×2 | 28 |
| 129 | 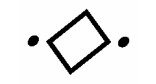 | mgbo | 2×1+4×2 | 4×2 | 18 |
| 130 | 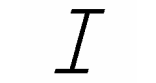 | na | 3×2 | 2×2 | 10 |
| 131 | 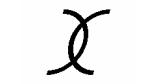 | nɛ | 2×3 | 2×3 | 12 |
| 132 | 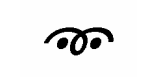 | ne | 2×1+3×3 | 2×1+1×3 | 16 |
| 133 | 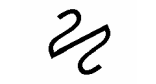 | ni | 1×2+4×3 | 2×1+2×2 | 20 |
| 134 | 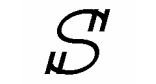 | nɔ | 4×2+4×3 | 3×1+2×2+2×3 | 33 |
| 135 | 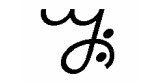 | no | 2×1+5×3 | 2×1+2×2+1×3 | 26 |
| 136 | 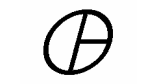 | nu | 2×2+2×3 | 2×1+4×2 | 20 |
| 137 | 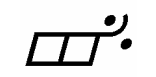 | nɗa | 2×1+4×2+1×3 | 6×2 | 25 |
| 138 | 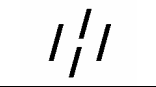 | nɗɛ | 4×2 | — | 8 |
| 139 | 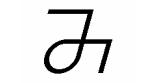 | nɗe | 2×2+2×3 | 1×1+2×2+1×3 | 18 |
| 140 | 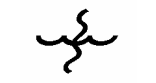 | nɗi | 8×3 | 4×1+2×2+1×3 | 35 |
| 141 | 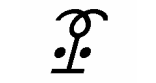 | nɗɔ | 2×1+2×2+3×3 | 2×1+2×2+1×3 | 24 |
| 142 | 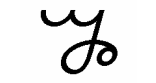 | nɗo | 6×3 | 3×1+3×2+1×3 | 30 |
| 143 | 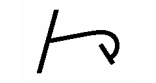 | nɗu | 2×2+1×3 | 2×2 | 11 |
| 144 | 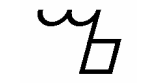 | ɲa | 4×2+2×3 | 6×2 | 26 |
| 145 | 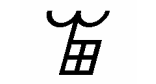 | ɲɛ | 6×2+2×3 | 11×2+1×3 | 43 |
| 146 | 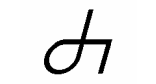 | ɲi | 1×2+2×3 | 1×1+1×2+1×3 | 14 |
| 147 | 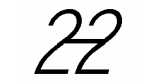 | ɲɔ | 3×2+6×3 | 4×1+4×2 | 36 |
| 148 | 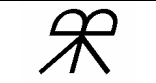 | ɲʤa | 3×2+2×3 | 11×2+1×3 | 37 |
| 149 | 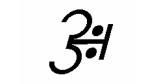 | ɲʤɛ | 2×1+3×2+4×3 | 2×1+4×2 | 30 |

| | sign | transliteration | components | connections | complexity |
|---|---|---|---|---|---|
| 150 | 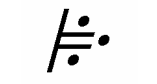 | ɲʤe | 3×1+3×2 | 2×2 | 13 |
| 151 |  | ɲʤi | 2×1+3×3 | 2×2 | 15 |
| 152 | 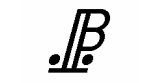 | ɲʤɔ | 2×1+3×2+2×3 | 8×2 | 30 |
| 153 | 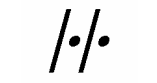 | ɲʤo | 2×1+2×2 | — | 6 |
| 154 | 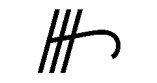 | ɲʤu | 3×2+1×3 | 1×2+2×3 | 17 |
| 155 | 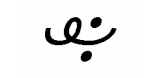 | ŋa | 2×1+3×3 | 2×1+1×3 | 16 |
| 156 | 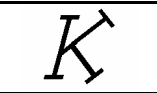 | ŋɛ | 7×2 | 6×2 | 26 |
| 157 | 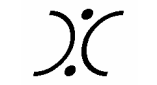 | ŋɔ | 2×1+2×3 | — | 8 |
| 158 | 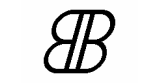 | ŋga | 2×2+4×3 | 2×1+10×2 | 38 |
| 159 | 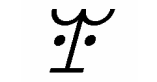 | ŋgɛ | 2×1+2×2+2×3 | 4×2 | 20 |
| 160 | 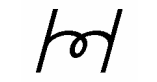 | ŋge | 2×2+3×3 | 2×1+2×2+1×3 | 22 |
| 161 | 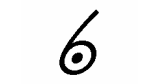 | ŋgi | 1×1+2×3 | 1×1+1×2 | 10 |
| 162 | 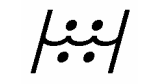 | ŋgɔ | 4×1+2×2+3×3 | 4×2 | 25 |
| 163 | 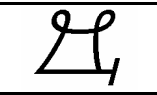 | ŋgo | 2×2+3×3 | 2×1+3×2+2×3 | 27 |
| 164 | 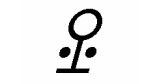 | ŋgu | 2×1+2×2+2×3 | 2×1+2×2 | 18 |
| 165 | 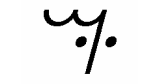 | pa | 2×1+1×2+2×3 | 2×2 | 14 |
| 166 | 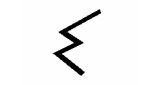 | pɛ | 4×2 | 3×2 | 14 |
| 167 | 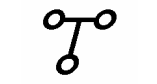 | pe | 2×2+6×3 | 6×1+4×2 | 36 |
| 168 | 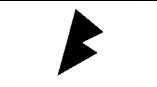 | pi | 5×2+2×2 | 7×2 | 28 |
| 169 | 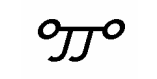 | pɔ | 1×2+6×3 | 4×1+4×2 | 32 |
| 170 | 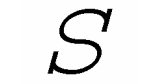 | po | 2×2+4×3 | 3×1+2×2 | 23 |
| 171 | 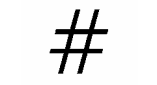 | pu | 4×2 | 4×3 | 20 |
| 172 | 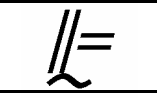 | ra | 4×2+2×3 | 1×1 | 15 |
| 173 | 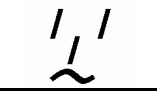 | rɛ | 3×2+2×3 | 1×1 | 13 |
| 174 | 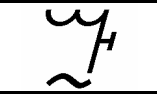 | re | 3×2+4×3 | 1×1+4×2 | 27 |
| 175 | 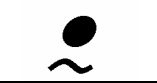 | ri | 4×3+1×2 | 3×1 | 17 |
| 176 | 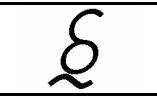 | rɔ | 5×3 | 3×1+1×2 | 20 |
| 177 | 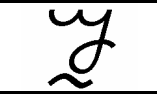 | ro | 6×3 | 2×1+2×2+1×3 | 27 |
| 178 | 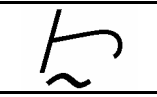 | ru | 1×2+3×3 | 1×1+1×2 | 14 |
| 179 | 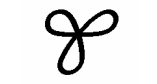 | sa | 6×3 | 6×1+3×3 | 33 |
| 180 | 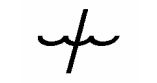 | sɛ | 1×2+3×3 | 2×2+1×3 | 18 |
| 181 | 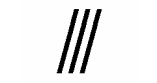 | se | 3×2 | — | 6 |
| 182 | 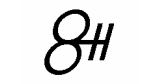 | si | 3×2+4×3 | 4×1+4×2+1×3 | 33 |
| 183 | 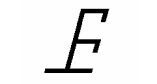 | sɔ | 6×2 | 5×2 | 22 |
| 184 | 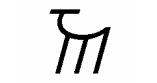 | so | 4×2+1×3 | 4×2 | 19 |
| 185 | 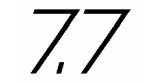 | su | 5×2 | 2×2 | 14 |
| 186 | 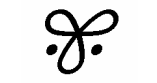 | ʃa | 2×1+6×3 | 6×1+3×3 | 35 |
| 187 | 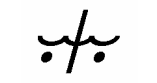 | ʃɛ | 2×1+1×2+3×3 | 2×2+1×3 | 20 |
| 188 | 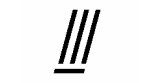 | ʃe | 4×2 | — | 8 |
| 189 | 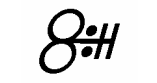 | ʃi | 2×1+3×2+4×3 | 4×1+4×2+1×3 | 35 |

| | sign | transliteration | components | connections | complexity |
|---|---|---|---|---|---|
| 190 | 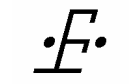 | ʃɔ | 2×1+6×2 | 5×2 | 24 |
| 191 | 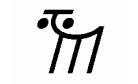 | ʃo | 2×1+4×2+1×3 | 4×2 | 21 |
| 192 | 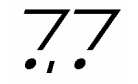 | ʃu | 2×1+5×2 | 2×2 | 16 |
| 193 | 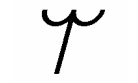 | ta | 1×2+2×3 | 3×2 | 14 |
| 194 | 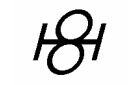 | tɛ | 3×2+4×3 | 4×1+5×2 | 32 |
| 195 | 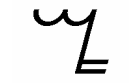 | te | 3×2+2×3 | 4×2 | 20 |
| 196 | 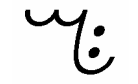 | ti | 2×1+3×3 | 2×2 | 15 |
| 197 | 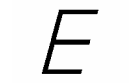 | tɔ | 4×2 | 3×2 | 14 |
| 198 | 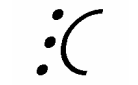 | to | 3×1+1×3 | — | 6 |
| 199 | 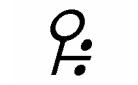 | tu | 2×1+2×2+2×3 | 2×1+2×2 | 18 |
| 200 | 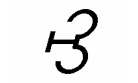 | va | 3×2+4×3 | 2×1+4×2 | 28 |
| 201 | 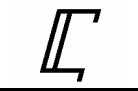 | vɛ | 6×2 | 6×2 | 24 |
| 202 | 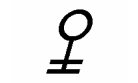 | ve | 3×2+2×3 | 2×1+2×2+1×3 | 21 |
| 203 | 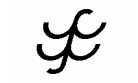 | vi | 6×3 | 1×1+6×2 | 31 |
| 204 | 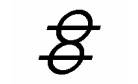 | vɔ | 2×2+4×3 | 4×1+1×2+4×3 | 34 |
| 205 | 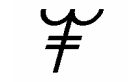 | vo | 3×2+2×3 | 3×2+2×3 | 24 |
| 206 | 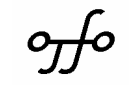 | vu | 1×2+7×3 | 5×1+5×2 | 38 |
| 207 | 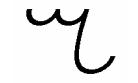 | wa | 3×3 | 2×2 | 13 |
| 208 | 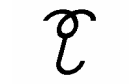 | wɛ | 4×3 | 2×1+1×2+1×3 | 19 |
| 209 | 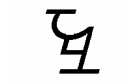 | we | 5×2+1×3 | 6×2 | 25 |
| 210 | 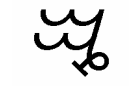 | wi | 1×2+6×3 | 1×1+5×2+1×3 | 34 |
| 211 | 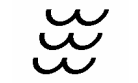 | wɔ | 6×3 | 3×2 | 24 |
| 212 | 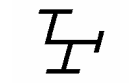 | wo | 6×2 | 5×2 | 22 |
| 213 | 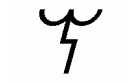 | wu | 3×2+2×3 | 5×2 | 22 |
| 214 | 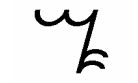 | wã | 1×2+4×3 | 4×2 | 22 |
| 215 | 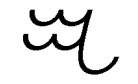 | ja | 5×3 | 4×2 | 23 |
| 216 | 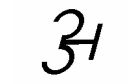 | jɛ | 3×2+4×3 | 2×1+4×2 | 28 |
| 217 | 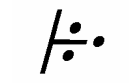 | je | 3×1+2×2 | 1×2 | 9 |
| 218 | 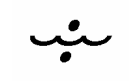 | ji | 2×1+3×3 | 2×2 | 15 |
| 219 | 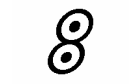 | jɔ | 2×1+4×3 | 4×1+1×2 | 20 |
| 220 | 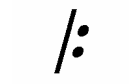 | jo | 2×1+1×2 | — | 4 |
| 221 | 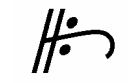 | ju | 2×1+2×2+1×3 | 1×2+1×3 | 14 |
| 222 | 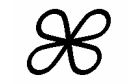 | za | 8×3 | 8×1+2×2+4×3 | 48 |
| 223 | 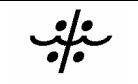 | zɛ | 4×1+1×2+3×3 | 2×2+1×3 | 22 |
| 224 | 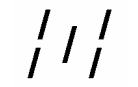 | ze | 5×2 | — | 10 |
| 225 | 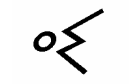 | zi | 4×2+2×3 | 2×1+3×2 | 22 |
| 226 | 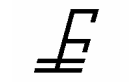 | zɔ | 7×2 | 5×2+1×3 | 27 |
| 227 | 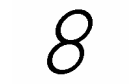 | zo | 4×3 | 4×1+1×2 | 18 |
| 228 | 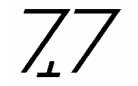 | zu | 6×2 | 3×2 | 18 |
| 229 | 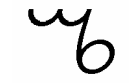 | ŋ | 4×3 | 1×1+3×2 | 19 |

* Bold numbers (**2**) correspond to the filled areas.

The distribution of complexities in all previously investigated scripts was uniform. Surprisingly, the hypothesis is rejected for the Vai script (cf. the following table).

Table 2

Distribution of complexities

| **C** | $f_C$ | **C** | $f_C$ | **C** | $f_C$ | **C** | $f_C$ | **C** | $f_C$ | **C** | $f_C$ | **C** | $f_C$ | **C** | $f_C$ | **C** | $f_C$ |
|---|---|---|---|---|---|---|---|---|---|---|---|---|---|---|---|---|---|
| **4** | 2 | **9** | 4 | **14** | 12 | **19** | 8 | **24** | 13 | **29** | 2 | **34** | 5 | **39** | 1 | **44** | 0 |
| **5** | 1 | **10** | 4 | **15** | 6 | **20** | 12 | **25** | 8 | **30** | 6 | **35** | 3 | **40** | 0 | **45** | 1 |
| **6** | 7 | **11** | 5 | **16** | 9 | **21** | 5 | **26** | 13 | **31** | 1 | **36** | 5 | **41** | 0 | **46** | 1 |
| **7** | 3 | **12** | 7 | **17** | 7 | **22** | 10 | **27** | 9 | **32** | 4 | **37** | 1 | **42** | 1 | **47** | 0 |
| **8** | 7 | **13** | 8 | **18** | 12 | **23** | 7 | **28** | 10 | **33** | 3 | **38** | 3 | **43** | 3 | **48** | 1 |

The uniformity hypothesis will be tested by the run test. Denote $I$ the inventory size and $R$ the range of complexities (for the Vai script we have $I = 229$ and $R = 44$). If the data are uniformly distributed, all expected frequency values are $E = \frac{I}{R+1}$. A run is a sequence of frequencies which are either all greater than $E$ or all smaller than $E$. Hence we have $E = \frac{229}{44+1} = 5.09$ and 11 runs, namely [2,1, 7, 3, 7, 4,4,5, 7,8,12,6,9,7,12,8,12, 5, 10,7,13,8,13,9,10, 2, 6, 1,4,3,5,3,5,1,3,1,0,0,1,2,0,1,1,0,1]. Denote $n = R + 1$, $n_1$ the number of frequencies smaller than $E$ and $n_2$ the number of frequencies greater than $E$ (in this case $n = 45$, $n_1 = 26$, $n_2 = 19$). The number of runs is considered random (and, consequently, the distribution is considered uniform) if

$$z = \frac{|r - E(r)| - 0.5}{\sigma_r} < 1.96,$$

where $r$ is the number of runs, $E(r) = 1 + \frac{2n_1 n_2}{n}$ and $\sigma_r = \sqrt{\frac{2n_1 n_2 (2n_1 n_2 - n)}{n^2 (n-1)}}$. We obtain $z = 3.55$, which means the Vai script is the first case where the uniform distribution does not yield a good fit.

Here, it is necessary to note the peculiarity of syllabic scripts in comparisons with alphabets. With the number of characters significantly higher, syllabaries usually contain some redundant signs as a result of additional filling in the gaps not occurring in original versions, cf. also syllable representations in the latest version of the Bamum script, another indigenous African invention (Schmitt 1963: Tab. 15; Mafundikwa 2004: 87–88). For the Vai script, the typical number of utilizable syllables hardly reaches a hundred (Singler 1996). That is, it would be interesting to check the uniformity hypothesis having sufficiently long native Vai texts to separate the core of the syllabary and its marginal part. This task together with character frequency will be addressed in future works.

In previous works (Buk, Mačutek and Rovenchak 2008, Mačutek 2008) the numbers of components and connections were also investigated (for Latin, Cyrillic, and Runic scripts). The Poisson distribution ($P_x = e^{-\lambda}\lambda^x / x!$, $\lambda > 0$) was applied as a model in both cases. The parameter $\lambda$ is the mean of the distribution, which leads to a quite straightforward interpretation – the numbers of components and connections are controlled by the respective means, a character with 'too many' components or connections occurs with a low probability. Moreover, as the parameter is also the variance of the distribution, the higher the mean, the higher variability is expected. A relatively high number of Vai characters without a connection makes it necessary to modify the respective model, the result being the hyper-Poisson distribution ($P_x = a^x / {}_1F_1(1; b; a)\, b^{(x)}$, $a \geq 0$, $b > 0$, ${}_1F_1(1; b; a)$ is a hypergeometric function). We remind that the Poisson distribution is its special case for $b = 1$, cf. Wimmer and Altmann (1999). The Vai script with its 229 characters provides another corroboration of the models.

Table 3

Numbers of components and connections

| | **components** | **connections** |
|---|---|---|
| 0 | | 17 |
| 1 | | 9 |
| 2 | 7 | 30 |
| 3 | 22 | 24 |
| 4 | 41 | 33 |
| 5 | 49 | 33 |
| 6 | 48 | 34 |
| 7 | 35 | 16 |
| 8 | 17 | 10 |
| 9 | 8 | 10 |
| 10 | 2 | 5 |
| 11 | | 3 |
| 12 | | 4 |
| 13 | | 0 |
| 14 | | 1 |
| | Poisson<br>$\lambda = 3.50$, $\chi^2 = 4.39$<br>$P = 0.73$, $DF = 7$ | Hyper-Poisson<br>$a = 10.73$, $b = 7.50$<br>$\chi^2 = 18.86$<br>$P = 0.09$, $DF = 12$ |

**Acknowledgement**

J. Mačutek was supported by the research grant VEGA 1/3016/06.

**References**

**Altmann, G.** (2004). Script complexity. *Glottometrics 8, 68-74.*

**Altmann, G.** (2008). Towards a theory of script. In: Altmann, G., Fan, F. (eds.), *Analyses of Script. Properties of Characters and Writing Systems: 149-164.* Berlin: de Gruyter.

**Bird, S.** (1999). Strategies for Representing Tone in African Writing Systems. *Written Language and Literacy 2, 1–44.*
**Buk, S., Mačutek, J., Rovenchak, A.** (2008). Some properties of the Ukrainian writing system. *Glottometrics 16, 63-79.*
**Census** (2008). *2008 National Population and Housing Census. Preliminary results.* Monrovia: Liberia Institute of Statistics and Geo-Information Services.
**Coulmas, F.** (2004). *The Blackwell Encyclopedia of Writing Systems*. Blackwell Publishing.
**Dalby, D.** (1967). A survey of the indigenous scripts of Liberia and Sierra Leone: Vai, Mende, Loma, Kpelle and Bassa. *African Language Studies 8, 1–51.*
**Everson, M., Nyei, M., Riley, Ch., Sherman, T.** (2006). Proposal for addition of Vai characters to the UCS. http://www.dkuug.dk/jtc1/sc2/wg2/docs/n3081.pdf.
**Galtier, G.** (1987): Un exemple d'écriture traditionnelle mandingue: le "Masaba" des Bambara-Masasi du Mali. *Journal des Africanistes 57(1), 255–266.*
**Gordon, R. G., Jr.** (ed.) (2005). Ethnologue: Languages of the World, Fifteenth edition. Dallas, Tex.: SIL International. Online version: http://www.ethnologue.com.
**Jensen, H.** (1969). *Die Schrift in Vergangenheit und Gegenwart. 3., neubearb. und erw. Aufl.* Berlin: Deutscher Verl. der Wissenschaften.
**Köhler, R.** (2008). Quantitative analysis of writing systems: An introduction. In: Altmann, G., Fan, F. (eds.), *Analyses of Script. Properties of Characters and Writing Systems: 3-9.* Berlin: de Gruyter.
**Mačutek, J.** (2008). Runes: complexity and distinctivity. *Glottometrics 16, 1-16.*
**Mafundikwa, S.** (2004). *Afrikan Alphabets: The Story of Writing in Afrika*. New York: Mark Butty Publisher.
**Massaquoi, M.** (1911). The Vai people and their syllabic writing. *Journal of the Royal African Society 10(40), 459–466.*
**Ofri-Scheps, D.** (1991). Vai Phonemics. In: *On the Object of Ethnology: Apropos of Vai Culture in Liberia: 84–119.* Ph. D. Dissertation, University of Berne.
**Priest, L. A.** (2004). Vai Syllabary. http://scripts.sil.org/cms/scripts/render_download.php?site_id=nrsi&format=file&media_id=VaiUnicode&filename=Vai+Syllabary2004-12-01.pdf.
**Schmitt, A.** (1963). *Die Bamum-Schrift: 3 Bde.* Wiesbaden: Otto Harrassowitz.
**Singler, J. V.** (1983). [Review of] The Psychology of Literacy. By Sylvia Scribner and Michael Cole. *Language 59(4), 893–901.*
**Singler, J. V.** (1996). Scripts of West Africa. In: Daniels, P. T., Bright, W. (eds.), *The World's Writing Systems: 593–598*. Oxford: Oxford University Press.
**Tuchscherer, K.** (2005). History of writing in Africa. In: Apiah, K. A., Gates, H. L., Jr. (eds.), *Africana: The Encyclopedia of the African and African American Experience (second edition): 476–480.* New York: Oxford University Press.
**Tuchscherer, K.** (2007). Recording, communicating, and making visible: A history of writing and systems of graphic symbolism in Africa. In: *Inscribing Meaning. Writing and Graphic Systems in African Art: 37–53*. Smithsonian.
**Tuchscherer, K., Hair, P. E. H.** (2002). Cherokee and West Africa: Examining the origins of the Vai script. *History in Africa 29, 427-486.*
**Tucker, S. L.** (1999). [illegible] *[Vai kpolo lɔ kowae gbu kpaha sɔ kpoloe fɔlanaã mɛ = Vai script primer 2]*. Monrovia: Libtralo.
**Vydrine, V.** (1999). *Manding–English Dictionary (Maninka, Bamana). Vol. 1.* St. Petersburg: Dimitry Bulanin Publishing House.
**Welmers, W. E.** (1976). *A Grammar of Vai*. Berkeley: University of California Press.

**Wimmer, G., Altmann, G.** (1999). *Thesaurus of Univariate Discrete Probability Distributions*. Essen: Stamm.